%% file: wsasr_icapssp27.tex
\documentclass{article}
\usepackage{spconf,amsmath,graphicx,hyperref}

\usepackage{amsfonts,amssymb}
\usepackage{array}
\usepackage{booktabs}
\usepackage{textcomp}
\usepackage{stfloats}
\usepackage{float}
\usepackage{url}
\usepackage{cite}
\usepackage{tikz}
\usepackage{enumitem}
\usetikzlibrary{arrows.meta}
\newcommand{\OTCw}{OTC-W$_{e}$}
\newcommand{\OTCt}{OTC-T$_{e}$}
\newcommand{\OTCr}{OTC-T$_{H}$}
\newcommand{\OTCtw}{OTC-TW$_{e}$}
\newcommand{\OTCtwr}{OTC-TW$_{H}$}
\newcommand{\OTCwt}{OTC-W$_{e}'$}

\newcommand{\OTCwg}{OTC-W}
\newcommand{\OTCtg}{OTC-T}
\newcommand{\OTCtwg}{OTC-TW}

\title{A TRAINING CRITERION WITH TOKEN-LEVEL TOLERANCE TO TRANSCRIPTION AMBIGUITY FOR AUTOMATIC SPEECH RECOGNITION}

\name{Saurabh Kumar, Diptiman Mohanta, Prasanta Kumar Ghosh}
\address{Department of Electrical Engineering, Indian Institute of Science (IISc), Bangalore, India\\
\{saurabhk0317, diptimanmohanta7\}@gmail.com, prasantg@iisc.ac.in}

\begin{document}
\ninept
\maketitle

\begin{abstract}
Automatic speech recognition is typically trained assuming that the reference transcript is the only valid labeling of an utterance, yet even nominally verbatim transcripts contain localized differences in pronunciation, spelling, or lexical realization that the acoustics do not uniquely determine. Omni-temporal Classification (OTC) tolerates such noise by adding wildcard paths to the connectionist temporal classification (CTC) alignment graph, but its word-level arcs are too coarse, since bypassing one unsupported token discards supervision for the whole word. We move wildcard arcs to token granularity so unsupported tokens can be bypassed while the rest of the word stays supervised, and we combine token- and word-level arcs as complementary escape paths. Across 19 languages and three corpora, token-level OTC improves over CTC on all 25 tasks. We also replace epoch-indexed relaxation of the wildcard weights with a predictive-entropy-indexed schedule, which performs comparably while reducing dependence on training length. Combining this schedule with the hybrid graph gives the lowest mean word error rate (WER) on every corpus and a 9.45\% average relative WER reduction over CTC. Independent validator transcriptions show that token-level models place significantly more wildcard-bypass probability than CTC on disputed characters, indicating that token-level tolerance targets localized transcript ambiguity.
\end{abstract}


\section{Introduction}
\label{sec:intro}

End-to-end automatic speech recognition (ASR) models trained with connectionist temporal classification (CTC) \cite{ctc2006} sum over alignments that collapse to a single reference transcript, treating it as the only valid labeling. Yet even verbatim transcripts contain choices the acoustics do not settle, such as spelling variants, vowel-length or nasalization marks in Indic scripts, or inflections reduced in fast speech. Annotators often disagree on one or two characters of a word while agreeing on the rest, which ASR evaluation increasingly accounts for through multiple references or permissible spelling variants \cite{ali2015multi,mcnamara2024style,bhogale2026oiwer}. CTC training, however, still supervises every position as if the reference were certain, pushing the model toward labels the audio may not support.

Error-tolerant objectives for CTC and transducers \cite{rnnt2012,wtransducer2023,wst2025} address several kinds of label noise. STC \cite{stc2022} and W-CTC \cite{wctc2022} handle missing parts of the transcript, graph-based temporal classification \cite{moritz2021semi} and alternative pseudo-labeling \cite{apl2023} handle uncertain pseudo-labels, and token-weighted RNN-T \cite{tokenweighted2024} down-weights unreliable tokens. Bypass temporal classification \cite{btc2023} and its extension, Omni-temporal Classification (OTC) \cite{otc2023}, add a wildcard $\star$ to the weighted finite-state transducer (WFST) alignment graph \cite{wfst2008,hannun2020differentiable}, whose bypass arcs skip unsupported transcript units and self-loops absorb unexplained frames; similar arcs between adjacent tokens also make keyword spotting robust to noise \cite{xi2024ntckws}. These methods mostly target heavy label corruption or acoustic noise, whereas we focus on verbatim transcripts, where the reference is largely correct and disagreement is sparse and local.

Published OTC uses word-level arcs with weights tuned on transcripts containing 50\% simulated errors and performs similarly to CTC on verbatim data \cite{otc2023}. To our knowledge, no OTC-related work reports a gain over CTC on verbatim LibriSpeech. Re-tuning the weights on clean speech helps only slightly since skipping one unsupported token still discards supervision for the whole word. We therefore place wildcard arcs at the token level so that a local deviation can be bypassed while the remaining tokens stay supervised, and we add word-level bypasses as a second escape path for wholly unsupported words. We also relax the wildcard weights by using held-out predictive entropy rather than epoch count to reduce dependence on training length.

WER gains alone do not show that this tolerance is used where transcripts are actually ambiguous. We therefore obtained independent transcriptions of the RESPIN \cite{kumar2026respin} development and test sets, for which only a single reference is released, from three external validation vendors. Disputed characters and words receive substantially more bypass probability than agreed ones, and token-level models assign significantly more of it than CTC to disputed characters while leaving agreed characters nearly unchanged. This indicates that token-level tolerance is associated with localised transcript ambiguity rather than an indiscriminate relaxation of the objective.

Our contributions can be summarised as follows.
\begin{itemize}
    \item Token-level wildcard arcs for OTC-based ASR training, which improve WER over CTC on all 25 tasks across 19 languages and three corpora, whereas existing word-level OTC performs similarly to CTC on LibriSpeech and RESPIN.
    \item A hybrid graph combining token- and word-level bypasses, trained with an entropy-indexed relaxation schedule, which gives the lowest mean WER on every corpus and a 9.45\% average relative reduction over CTC.
    \item An analysis based on independent validator transcriptions of RESPIN, showing that token-level models become more tolerant specifically at characters where annotators disagree, which links the added tolerance to genuine transcript ambiguity in verbatim data.\footnote{Code and configurations are available at
\url{https://github.com/saurabhk0317/wsasr_icassp27.git}.}
\end{itemize}
\section{Background}
\label{sec:background}

Let $\mathbf{y}=[y_1,\ldots,y_U]$ be a transcript over vocabulary $\mathcal{V}$ including the CTC blank $\varnothing$, and let $P_t(v)$ be the model posterior of $v\in\mathcal{V}$ at frame $t$ of a $T$-frame utterance. CTC marginalises over the frame-level alignments that collapse to $\mathbf{y}$, obtained by composing the CTC topology with a linear acceptor of $\mathbf{y}$~\cite{ctc2006,wfst2008,hannun2020differentiable}, so every valid path emits the transcript exactly (Fig.~\ref{fig:graphs}a).

\subsection{Omni-temporal Classification}
\label{sec:otc}

OTC~\cite{otc2023} adds a wildcard $\star$ to this graph. For a segmentation $\mathbf{y}=u_1\circ\cdots\circ u_K$ with states $0,\ldots,K$, states $k-1$ and $k$ are joined by an arc emitting $u_k$ and a parallel \emph{bypass} arc labelled $\star$ of log-weight $w^{(\mathrm{byp})}$, and every state carries a $\star$ \emph{self-loop} of log-weight $w^{(\mathrm{self})}$; wildcard emissions are scored by the mean non-blank posterior~\cite{stc2022}. Let $\Pi(\mathbf{y})$ denote the alignments admitted by this graph composed with the CTC topology, and let $n_{\mathrm{byp}}(\pi)$ and $n_{\mathrm{self}}(\pi)$ count bypass and self-loop arcs on a path $\pi$, giving the wildcard-arc cost $\phi(\pi)=n_{\mathrm{byp}}(\pi)w^{(\mathrm{byp})}+n_{\mathrm{self}}(\pi)w^{(\mathrm{self})}$. The OTC objective is
\begin{equation}
\label{eq:otc}
\mathcal{L}_{\mathrm{OTC}}
=
-\log\sum_{\pi\in\Pi(\mathbf{y})}
\left(\prod_{t=1}^{T}P_t(\pi_t)\right)e^{\phi(\pi)},
\end{equation}
which reduces to CTC as both wildcard weights tend to $-\infty$ and wildcard paths are suppressed. The weights start strongly negative, so the transcript is trusted early in training, and are relaxed geometrically with epoch $e$:
\begin{equation}
\label{eq:decay}
w^{(j)}(e)=w^{(j)}_1\,\tau_j^{\,e-1},
\quad
j\in\{\mathrm{byp},\mathrm{self}\},\quad \tau_j\in(0,1).
\end{equation}
In~\cite{otc2023} each $u_k$ is a word, so a bypass spans a whole word and self-loops occur only at word boundaries (Fig.~\ref{fig:graphs}b). More generally, the segmentation determines the wildcard granularity, which we move to token level in Sec.~\ref{sec:granularity}.

\section{Method}
\label{sec:method}
\input{figures/transcript_graphs}

\subsection{Arc granularity}
\label{sec:granularity}

We target localized transcript variation, where a word differs from the acoustics in only one or a few tokens because of pronunciation, spelling, or inflectional variation, which motivates placing wildcard arcs at token rather than word granularity. Whereas the published \OTCw{} formulation treats each word as one unit, \OTCt{} sets $K=U$, placing a bypass arc between every pair of consecutive tokens and a self-loop at every state (Fig.~\ref{fig:graphs}c); word boundaries are obtained from the SentencePiece word-start marker for BPE units and from the space symbol for character units.

Token-level arcs provide finer resolution and retain supervision: a single unsupported token can be bypassed without skipping the whole word, leaving the remaining tokens on the alignment path. They also make the escape cost less dependent on word length, since \OTCwg{} replaces all $n$ token emissions of a bypassed word with one $\star$ at a fixed penalty, whereas token-level arcs incur a cost that scales with the number of bypassed tokens. The two are therefore complementary: a wholly unsupported word can be handled by one word-level bypass, and a partially unsupported word by token-level bypasses. We combine both in \OTCtwg{}, which adds one word bypass arc per word to the token-level graph (Fig.~\ref{fig:graphs}d). The self-loop absorbs extra audio frames and doesn't need a separate word-level version because the token-level self-loop can already absorb any number of frames.

\subsection{Entropy-indexed decay}
\label{sec:recip}

The geometric decay in Eq.~\eqref{eq:decay} uses epoch as a proxy for training progress, so a schedule tuned for one training budget need not reach the same learning stage under another. We instead measure progress by predictive entropy on held-out data, computed without gradients so that it does not reflect memorization of the training transcripts. At each epoch, we remove the CTC blank and compute the normalized non-blank posterior and its entropy at each frame:
\begin{equation}
\label{eq:q}
q_t(v)=\frac{P_t(v)}{1-P_t(\varnothing)},\qquad
H_t=-\sum_{v\neq\varnothing}q_t(v)\log q_t(v).
\end{equation}
Averaging with the non-blank mass $\omega_t=1-P_t(\varnothing)$ and normalizing by $\log(V-1)$, where $V=|\mathcal{V}|$, gives $\widehat{H}_e\in[0,1]$:
\begin{equation}
\label{eq:H}
\widehat{H}_e=
\frac{1}{\log(V-1)}
\frac{\sum_{t\in\mathcal{D}_{\mathrm{val}}}\omega_t H_t}
     {\sum_{t\in\mathcal{D}_{\mathrm{val}}}\omega_t}.
\end{equation}
Let $H_0$ be $\widehat{H}_e$ at the end of the first epoch. We smooth the entropy as $\bar H_e=\alpha\widehat H_e+(1-\alpha)\bar H_{e-1}$ and convert it to progress using the reciprocal $r(h)=1/(h+\epsilon)$:
\begin{equation}\vspace{-2mm}
\label{eq:progress}
p_e=
\min\!\left\{1,\max\!\left\{0,
\frac{r(\bar H_e)-r(H_0)}
     {r(H_{\mathrm{low}})-r(H_0)}
\right\}\right\}.
\end{equation}
With $\beta_e=p_e^{\,\kappa}$, the wildcard weights are interpolated geometrically between their initial and final magnitudes:
\begin{equation}
\label{eq:interp}
w^{(j)}(e)=
-\left|w^{(j)}_1\right|^{1-\beta_e}
 \left|w^{(j)}_\infty\right|^{\beta_e},
\quad j\in\{\mathrm{byp},\mathrm{self}\}.
\end{equation}
Relaxation thus follows model confidence rather than elapsed epochs. 

\section{Experiments and Results}\vspace{-2mm}
\label{sec:experiments}

We evaluate our methods on 19 languages from LibriSpeech~\cite{librispeech2015}, FLEURS~\cite{conneau2023fleurs}, and RESPIN~\cite{kumar2026respin}. LibriSpeech uses \textit{train-clean-100} for training, \textit{dev-clean/other} for validation, and \textit{test-clean/other} for evaluation. FLEURS covers 14 languages across four families, 14 scripts, and diverse morphological types using the published splits. RESPIN uses the official dev/test sets and a 30-hour subset of \texttt{\textit{train\_\textless lang\textgreater\_small}}, restricted to unique sentences. Language codes are listed in Table~\ref{tab:main}.

All systems use a 12-block E-Branchformer encoder~\cite{ebranchformer2023} ($d=256$, four heads) in ESPnet2~\cite{espnet2018}, with WFST losses computed in k2. RESPIN uses 80-dimensional filterbanks for 70 epochs, while FLEURS and LibriSpeech use frozen XLS-R 300M~\cite{DBLP:journals/corr/abs-2111-09296} and wav2vec~2.0 Base features~\cite{wav2vec2020}, respectively, followed by linear projection and 30 epochs. FLEURS/RESPIN use character units, while LibriSpeech uses 200 BPE units. All systems apply SpecAugment~\cite{specaugment2019}, three-way speed perturbation, Adam with warmup, and validation-loss checkpoint averaging; only the loss function varies within each corpus. Decoding uses CTC prefix search without an external language model, and we report WER. To test whether the results depend on the encoder, we repeat the LibriSpeech experiments with a 12-block Conformer~\cite{conformer2020} of the same width and head count, changing only the encoder and adding early stopping (patience 5).

\subsection{Training criteria}\vspace{-2mm}
\label{sec:baselines}
\input{tables/parameters}
All systems share the same graph construction and WFST forward--backward implementation, differing only in wildcard arcs and schedules (Table~\ref{tab:criteria}); CTC disables both arc types. \OTCw{} uses the published settings~\cite{otc2023}, raising $|w^{(\mathrm{byp})}_1|$ only for divergent runs; \OTCwt{} re-tunes them on 10\,h of LibriSpeech \textit{train-clean-100}, and token-level settings come from the sweep in Sec.~\ref{sec:res-sweep}. Combined systems reuse the \OTCwt{} word bypass untuned. The entropy-indexed systems use $\alpha=0.3$, $H_{\mathrm{low}}=0.01$, $\kappa=1.3$, and $\epsilon=10^{-4}$ throughout, relaxing toward $w^{(\mathrm{byp})}_\infty=-0.01$ and $w^{(\mathrm{self})}_\infty=-0.001$. In \OTCtwr{}, the word bypass follows the same $p_e$ toward $w^{(\mathrm{wbyp})}_\infty=-1$, so one entropy measurement controls all three arc types. The final weights remain negative, so escapes are never free, and rising held-out entropy makes the weights more restrictive.\vspace{-2mm}

\subsection{Overall results}\vspace{-2mm}
\label{sec:res-main}

\input{tables/main_results}
\input{tables/conformer_results}

Table~\ref{tab:main} compares the seven criteria on the 25 tasks. With the published settings, OTC-W$_e$ has a slightly higher mean WER than CTC on LibriSpeech and RESPIN, and five runs required a larger bypass penalty to avoid divergence.\footnote{This matches the near-parity with CTC on LibriSpeech clean data reported in \cite{otc2023}. Absolute WERs are not directly comparable due to different toolkits and front-ends.} Re-tuning on clean speech (OTC-W$'_e$) removes the divergence, but the average gain is small and does not carry over to RESPIN. Word-level arcs therefore remain limited on verbatim transcripts, since bypassing one token skips the whole word.

Token-level tolerance gives a substantially different result. Both \OTCt{} and entropy-indexed \OTCr{} improve over CTC on all 25 tasks without per-language tuning. Their similar performance across all three corpora also shows that the gain is not specific to the choice of epoch- or entropy-indexed decay. The consistent gains support our hypothesis that allowing individual unsupported tokens to be bypassed is better suited to localised transcript variation than bypassing whole words. These comparisons also bound how much of the gain can be attributed to generic relaxation of the objective. \OTCw{} and \OTCwt{} add wildcard escapes to the same objective, the latter with weights re-tuned on clean speech, yet every token-level criterion has a lower mean WER than both of them on all three corpora, and \OTCw{} is worse than CTC on two. Relaxing the alignment constraint is therefore not sufficient on its own; what separates the criteria is where the escape is placed, which Sec.~\ref{sec:bypass} examines directly.

The combined graph further shows that the two granularities are complementary. \OTCtw{} improves over the token-only and word-only criteria on LibriSpeech and FLEURS, but its epoch-indexed relaxation degrades on RESPIN, falling 0.24\% below CTC on average, the same failure mode seen in \OTCw{} and \OTCwt{}. Replacing the epoch index with predictive entropy in \OTCtwr{} removes this degradation: it gives the lowest mean WER on all three corpora and, at 9.45\% relative, the largest average reduction over CTC of any criterion in Table~\ref{tab:main}, ahead of \OTCt{} (8.46\%) and \OTCr{} (7.67\%). It also has the lowest WER on 9 of the 25 individual tasks, more than any other single criterion.

Table~\ref{tab:conformer} repeats the LibriSpeech comparison with the Conformer encoder. The ranking of Table~\ref{tab:main} is reproduced exactly except that \OTCw{} and CTC exchange places, indicating that the benefit of token-level tolerance is not specific to the E-Branchformer.

\subsection{Sensitivity to the schedule}
\label{sec:res-sweep}

\begin{figure}[t]
\centering
\includegraphics[width=\columnwidth]{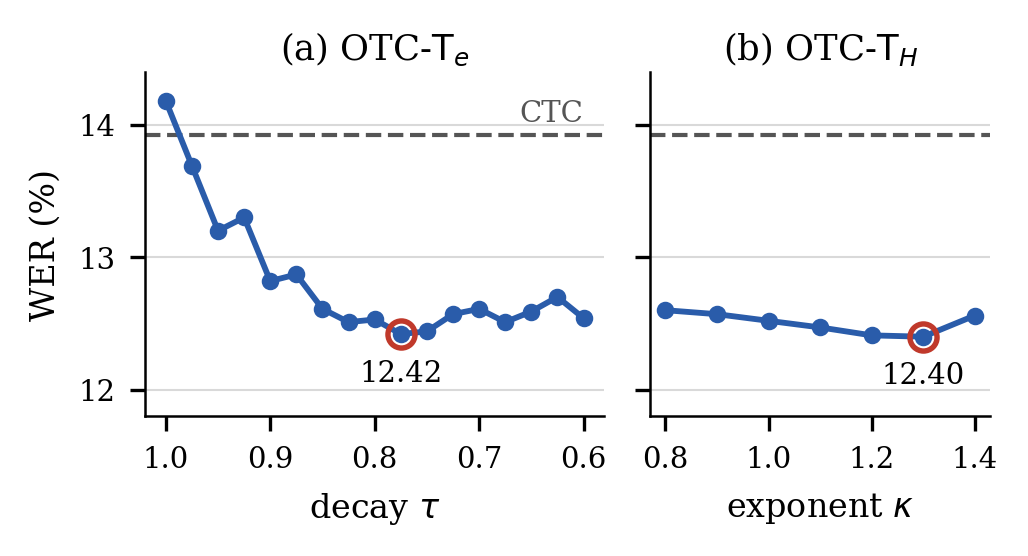}
\caption{Sensitivity of (a) \OTCt{} to the decay parameter $\tau$ and (b) \OTCr{} to the exponent $\kappa$ on LibriSpeech dev-clean+dev-other. WER (\%) is reported for models trained on a random 10\,h subset of train-clean-100 with the SSL front-end frozen. Dashed line: CTC; circled point: best setting.}

\label{fig:sweep}
\end{figure}

Figure~\ref{fig:sweep} examines the schedule parameters on a randomly selected 10\,h subset of LibriSpeech \texttt{train-clean-100}. Epoch-indexed \OTCt{} is sensitive to $\tau$, with overly slow relaxation performing worse than CTC and an optimum near $\tau=0.775$. In contrast, \OTCr{} is substantially less sensitive to $\kappa$ and improves over CTC throughout the tested range, with its best value at $\kappa=1.3$. We therefore use $\tau=0.775$ and $\kappa=1.3$ in the remaining experiments and select the \OTCwt{} word-level constants on the same subset. No per-task tuning is used in Table~\ref{tab:main}.

\subsection{Bypass probability and annotator disagreement}
\label{sec:bypass}
\input{tables/bypass_prob}

RESPIN releases one reference transcript per utterance~\cite{kumar2026respin}; we additionally obtained three independent transcriptions of its dev and test sets from external validation vendors (in-house, not yet public). Aligning each validator transcript to the reference at the character level, we call a character or word \emph{disputed} when at least one validator changes it \emph{and} the validators do not all propose the same alternative, since a unanimous correction indicates an erroneous rather than an ambiguous reference.

For each model we compose its frame posteriors with the wildcard transcript graph of the reference (Sec.~\ref{sec:otc}) and compute by forward--backward the posterior probability $m_k$ that an alignment takes the bypass arc at unit $k$,
\begin{equation}
\label{eq:bypassprob}
m_k=\frac{\sum_{\pi\in\Pi_k(\mathbf{y})}\left(\prod_{t}P_t(\pi_t)\right)e^{\phi(\pi)}}
         {\sum_{\pi\in\Pi(\mathbf{y})}\left(\prod_{t}P_t(\pi_t)\right)e^{\phi(\pi)}},
\end{equation}
where $\Pi_k(\mathbf{y})\subset\Pi(\mathbf{y})$ are the alignments that bypass $k$ and the denominator is the sum in Eq.~\eqref{eq:otc}. Thus $m_k$ reflects the training objective rather than the decoded output. All systems are probed at the same fixed weight ($w^{(\mathrm{byp})}=0$), so differences between rows of Table~\ref{tab:disagree} reflect the posteriors alone; CTC, never trained with wildcard arcs, is scored through the same graph as a control.

Every system, CTC included, places an order of magnitude more bypass probability on disputed than on agreed characters ($11\times$ for CTC), and the amount rises with the number of dissenting validators (Table~\ref{tab:disagree}). Token-level arcs raise the disputed-character mean by $27$--$36\%$ over CTC, with the combined criteria behaving alike, while agreed characters are essentially unchanged; the gain holds in all nine languages on both splits and is significant under a paired bootstrap over utterances ($10\,000$ resamples, $p<0.001$).

Word-level values are far smaller, and the criteria carrying word arcs raise agreed words as much as disputed ones, so their higher disputed means reflect a uniform increase rather than selectivity. We do not read much into these differences, since a whole word is rarely unsupported in verbatim transcripts, where half of all disputed words differ in a single character of an average $4.8$ and a third only by an insertion. This plausibly explains why word-level arcs help little here: a word bypass must discard supervision for the remaining characters to absorb a disagreement that a token arc handles far more cheaply, the mechanism hypothesised in Sec.~\ref{sec:granularity}.

\section{Conclusion}\vspace{-3mm}
\label{sec:conclusion}
Moving OTC's wildcard arcs from words to tokens lets an unsupported token be bypassed while the rest of the word stays supervised. With one setting for all corpora, token-level OTC improved over CTC on all 25 tasks, whereas published word-level OTC stayed close to CTC on LibriSpeech and RESPIN; the trends held with a Conformer encoder. Combining token- and word-level bypasses with entropy-indexed decay gave the lowest mean WER on every corpus, a $9.45\%$ average relative reduction over CTC, and avoided the RESPIN degradation of its epoch-indexed counterpart. Validator transcriptions show that disagreement is local, usually a single character or an inserted space, and token-level models assign more bypass probability to disputed characters while leaving agreed ones essentially unchanged. In future work, we will aim to learn the wildcard weights instead of scheduling them and evaluate against multiple references.

\section{Acknowledgments}
This work was partly supported by the RESPIN project, funded by the Gates Foundation. We thank the RESPIN contributors for data collection and validation. AI models (Claude Opus 5, Claude Sonnet 5, and GPT-5.6 Luna) assisted with writing, code, plots, tables, and references; all AI-assisted content was reviewed and verified by the authors.

\bibliographystyle{IEEEbib_etal}
\bibliography{refs}

\end{document}

%% file: figures/transcript_graphs.tex
\providecommand{\OTCtw}{OTC-TW$_{e}$}
\begin{figure}[t]
\centering
\begin{tikzpicture}[
  >={Stealth[length=1.1mm, width=0.85mm]},
  st/.style={circle, draw, minimum size=4.2mm, inner sep=0pt, line width=0.5pt, fill=white},
  fin/.style={st, double, double distance=0.45mm},
  lb/.style={font=\scriptsize, inner sep=1pt},
  wlabel/.style={font=\scriptsize\bfseries, text=gray!80},
  tag/.style={font=\scriptsize\bfseries, anchor=east},
  ctc/.style={->, draw=orange!90!black, line width=0.65pt},
  byp/.style={->, dashed, draw=blue!85!black, line width=0.65pt},
  slf/.style={->, densely dashed, draw=green!50!black, line width=0.6pt},
  wbyp/.style={->, dash dot, draw=blue!85!black, line width=0.75pt},
  wb/.style={draw=gray!50, densely dotted, line width=0.8pt},
  spanbar/.style={draw=gray!50, line width=0.5pt},
  x=1.15cm
]

\fill[blue!6, rounded corners=2pt] (-0.3, -5.9) rectangle (2.0, 0.5);
\fill[orange!6, rounded corners=2pt] (2.0, -5.9) rectangle (4.3, 0.5);
\draw[spanbar] (0.1, 0.65) -- (1.9, 0.65);
\node[wlabel] at (1, 0.85) {$w_1$};
\draw[spanbar] (2.1, 0.65) -- (3.9, 0.65);
\node[wlabel] at (3, 0.85) {$w_2$};
\draw[wb] (2, -6.0) -- (2, 1.0);

\begin{scope}[yshift=0mm]
  \node[tag] at (-0.45, 0) {(a) CTC};
  \foreach \i in {0,1,2,3} { \node[st] (a\i) at (\i, 0) {}; }
  \node[fin] (a4) at (4, 0) {};
  \foreach \i/\j/\y in {0/1/1, 1/2/2, 2/3/3, 3/4/4} {
    \draw[ctc] (a\i) -- (a\j) node[lb, midway, above=1pt] {$y_\y$};
  }
\end{scope}

\begin{scope}[yshift=-14mm]
  \node[tag] at (-0.45, 0) {(b) \OTCwg{}};
  \foreach \i in {0,1,2,3} { \node[st] (b\i) at (\i, 0) {}; }
  \node[fin] (b4) at (4, 0) {};
  \foreach \i/\j/\y in {0/1/1, 1/2/2, 2/3/3, 3/4/4} {
    \draw[ctc] (b\i) -- (b\j) node[lb, midway, above=1pt] {$y_\y$};
  }
  \draw[byp] (b0) to[out=48, in=132] node[lb, midway, above=-0.5pt] {$\star$} (b2);
  \draw[byp] (b2) to[out=48, in=132] node[lb, midway, above=-0.5pt] {$\star$} (b4);
  \foreach \i in {0,2,4} {
    \draw[slf] (b\i) to[out=240, in=300, looseness=6.5] node[lb, midway, below=1pt] {$\star$} (b\i);
  }
\end{scope}

\begin{scope}[yshift=-29mm]
  \node[tag] at (-0.45, 0) {(c) \OTCtg{}};
  \foreach \i in {0,1,2,3} { \node[st] (c\i) at (\i, 0) {}; }
  \node[fin] (c4) at (4, 0) {};
  \foreach \i/\j/\y in {0/1/1, 1/2/2, 2/3/3, 3/4/4} {
    \draw[ctc] (c\i) -- (c\j) node[lb, midway, above=1pt] {$y_\y$};
  }
  \foreach \i/\j in {0/1, 1/2, 2/3, 3/4} {
    \draw[byp] (c\i) to[out=60, in=120] node[lb, midway, above=-0.5pt] {$\star$} (c\j);
  }
  \foreach \i in {0,1,2,3,4} {
    \draw[slf] (c\i) to[out=240, in=300, looseness=6.5] node[lb, midway, below=1pt] {$\star$} (c\i);
  }
\end{scope}

\begin{scope}[yshift=-50mm]
  \node[tag] at (-0.45, 0) {(d) \OTCtwg{}};
  \foreach \i in {0,1,2,3} { \node[st] (d\i) at (\i, 0) {}; }
  \node[fin] (d4) at (4, 0) {};
  \foreach \i/\j/\y in {0/1/1, 1/2/2, 2/3/3, 3/4/4} {
    \draw[ctc] (d\i) -- (d\j) node[lb, midway, above=1pt] {$y_\y$};
  }
  \foreach \i/\j in {0/1, 1/2, 2/3, 3/4} {
    \draw[byp] (d\i) to[out=60, in=120] node[lb, midway, above=-0.5pt] {$\star$} (d\j);
  }
  \draw[wbyp] (d0) to[out=78, in=102, looseness=1.15]
    node[lb, midway, above=-0.5pt] {$\star$} (d2);
  \draw[wbyp] (d2) to[out=78, in=102, looseness=1.15]
    node[lb, midway, above=-0.5pt] {$\star$} (d4);
  \foreach \i in {0,1,2,3,4} {
    \draw[slf] (d\i) to[out=240, in=300, looseness=6.5] node[lb, midway, below=1pt] {$\star$} (d\i);
  }
\end{scope}

\begin{scope}[yshift=-62mm, xshift=0.10cm]
  \draw[ctc] (0, 0) -- (0.5, 0) node[lb, text=black, right=2pt] {CTC};
  \draw[byp] (1.05, 0) -- (1.55, 0) node[lb, text=black, right=2pt] {bypass};
  \draw[slf] (2.35, 0) -- (2.85, 0) node[lb, text=black, right=2pt] {self-loop};
\end{scope}

\end{tikzpicture}

\caption{Transcript graphs for words $w_1=(y_1,y_2)$ and $w_2=(y_3,y_4)$. (a) CTC. (b) \OTCwg{}: wildcard arcs at word boundaries. (c) \OTCtg{}: wildcard arcs at every token. (d) \OTCtwg{}: (c) plus one word-level bypass per word.}
\label{fig:graphs}
\end{figure}

%% file: tables/parameters.tex
\begin{table}[t]
\centering
\caption{Wildcard-arc settings. $w_1$: initial weight; $\tau$: per-epoch decay; E: entropy-indexed schedule (Sec.~\ref{sec:recip}).}
\label{tab:criteria}
\scriptsize
\setlength{\tabcolsep}{4pt}
\begin{tabular}{lcccccc}
\toprule
 & \multicolumn{2}{c}{Bypass} & \multicolumn{2}{c}{Self-loop} & \multicolumn{2}{c}{Word bypass} \\
\cmidrule(lr){2-3}\cmidrule(lr){4-5}\cmidrule(lr){6-7}
System & $w_1$ & $\tau$ & $w_1$ & $\tau$ & $w_1$ & $\tau$ \\
\midrule
\OTCw{}   & $-19$ & 0.975 & $3.75$ & 0.999 & --    & --    \\
\OTCwt{}  & $-19$ & 0.925 & $-1$   & 0.975 & --    & --    \\
\OTCt{}   & $-25$ & 0.775 & $-5$   & 0.775 & --    & --    \\
\OTCr{}   & $-25$ & E     & $-5$   & E     & --    & --    \\
\OTCtw{}  & $-25$ & 0.775 & $-5$   & 0.775 & $-19$ & 0.925 \\
\OTCtwr{} & $-25$ & E     & $-5$   & E     & $-19$ & E     \\
\bottomrule
\end{tabular}
\end{table}

%% file: tables/main_results.tex
\begin{table}[!t]
\caption{WER (\%) on 25 tasks; bold marks the row minimum. $W_e$, $W_e'$: word-level OTC (published, re-tuned weights); $T_e$, $T_H$: token-level OTC (epoch-, entropy-indexed); $TW_e$, $TW_H$: combined. $\Delta$: relative WER reduction (\%) of \OTCtwr{} vs.\ CTC. $\dag$: \OTCw{} diverged (see Sec.~\ref{sec:res-main}). Lavg, Favg, Ravg: mean WER over LibriSpeech, Fleurs, and RESPIN tasks; Avg: overall mean.}
\label{tab:main}
\centering
\scriptsize
\setlength{\tabcolsep}{2pt}
\begin{tabular}{lrrrrrrrr}
\toprule
\textbf{Lang.} & \textbf{CTC} & \textbf{W$_e$} & \textbf{W$_e'$} & \textbf{T$_e$}
& \textbf{T$_H$} & \textbf{TW$_e$} & \textbf{TW$_H$} & $\Delta$ \\
\midrule

\multicolumn{9}{l}{\textit{LibriSpeech}} \\
test-clean & 5.96 & 6.13 & 5.68 & 5.49 & 5.51 & \textbf{5.24} & 5.38 & 9.73 \\
test-other & 13.19 & 13.91 & 13.09 & 13.02 & 12.99 & 12.71 & \textbf{12.46} & 5.53 \\

\midrule

\multicolumn{9}{l}{\textit{FLEURS}} \\
ar\_eg & 33.96 & 35.70 & 34.42 & 30.61 & 31.38 & \textbf{30.15} & 30.78 & 9.36 \\
hy\_am & 26.56 & 23.84 & 25.36 & 23.25 & \textbf{22.50} & 22.97 & 22.89 & 13.82 \\
bn\_in & 35.31 & 33.60 & 35.07 & 32.24 & \textbf{31.14} & 31.76 & 31.75 & 10.08 \\
en\_us & 33.83 & 32.32 & 31.35 & 31.35 & 31.83 & 30.74 & \textbf{30.65} & 9.40 \\
ka\_ge & 48.78 & 45.75 & 47.51 & 43.18 & 43.53 & 43.80 & \textbf{43.07} & 11.71 \\
el\_gr & 39.08 & \textbf{30.44} & 33.05 & 33.35 & 36.31 & 32.52 & 32.06 & 17.96 \\
gu\_in & 36.49 & 34.24 & 34.42 & 31.92 & \textbf{31.68} & 32.04 & 31.85 & 12.72 \\
hi\_in & 31.71 & 32.33 & 31.91 & 28.40 & 27.80 & \textbf{27.40} & 27.69 & 12.68 \\
kn\_in & 35.87 & 33.77 & 35.62 & 31.96 & 32.58 & 31.97 & \textbf{31.93} & 10.98 \\
ml\_in & 37.48 & 35.71$^{\dag}$ & 37.19 & 34.81 & 34.74 & \textbf{34.38} & 34.57 & 7.76 \\
pa\_in & 39.56 & 42.68 & 39.18 & 36.67 & 36.99 & \textbf{36.25} & 36.89 & 6.75 \\
ru\_ru & 42.05 & 38.53 & 40.96 & 37.48 & 36.94 & 36.77 & \textbf{35.91} & 14.60 \\
ta\_in & 54.76 & 52.29$^{\dag}$ & 53.66 & 50.72 & 54.66 & 51.08 & \textbf{50.23} & 8.27 \\
te\_in & 48.01 & 44.76 & 46.65 & 42.69 & 42.61 & 42.06 & \textbf{41.93} & 12.66 \\

\midrule

\multicolumn{9}{l}{\textit{RESPIN}} \\
bh & 20.74 & 20.31 & 20.46 & 19.87 & 19.98 & 20.05 & \textbf{19.59} & 5.54 \\
bn & 22.02 & 21.75 & 23.00 & 20.94 & \textbf{20.14} & 22.02 & 20.18 & 8.36 \\
ch & 14.43 & 14.93 & 14.42 & \textbf{13.70} & 13.96 & 13.98 & 13.76 & 4.64 \\
hi & 15.69 & 15.33 & 15.45 & \textbf{13.62} & 13.98 & 14.81 & 14.03 & 10.58 \\
kn & 30.96 & 31.14$^{\dag}$ & 32.69 & \textbf{29.77} & 30.34 & 32.11 & 30.32 & 2.07 \\
mg & 26.70 & 26.92 & 26.57 & 25.94 & 26.39 & 26.18 & \textbf{25.65} & 3.93 \\
mr & 20.00 & 20.44$^{\dag}$ & 22.84 & 18.78 & \textbf{18.61} & 20.76 & 19.03 & 4.85 \\
mt & 23.42 & 23.89 & 23.11 & 22.38 & \textbf{21.88} & 22.76 & 22.03 & 5.94 \\
te & 27.39 & 27.73$^{\dag}$ & 30.51 & 27.20 & \textbf{26.87} & 29.16 & 27.15 & 0.88 \\

\midrule

\textbf{Lavg} & 9.58 & 10.02 & 9.39 & 9.26 & 9.25 & 8.98 & \textbf{8.92} & 6.84 \\
\textbf{Favg} & 38.82 & 36.85 & 37.60 & 34.90 & 35.34 & 34.56 & \textbf{34.44} & 11.27 \\
\textbf{Ravg} & 22.37 & 22.49 & 23.23 & 21.36 & 21.35 & 22.43 & \textbf{21.30} & 4.77 \\
\textbf{Avg} & 30.56 & 29.54 & 30.17 & 27.97 & 28.21 & 28.15 & \textbf{27.67} & 9.45 \\

\bottomrule
\end{tabular}\vspace{-2mm}
\end{table}

%% file: tables/conformer_results.tex
\begin{table}[t]
\centering
\caption{Conformer WERs (\%) on LibriSpeech for all methods reported in Table~\ref{tab:main}. $\Delta$: relative WER reduction over CTC (\%).}
\label{tab:conformer}
\scriptsize
\setlength{\tabcolsep}{3pt}
\begin{tabular}{lrrrrrrr}
\toprule
 & \textbf{CTC} & \textbf{W$_e$} & \textbf{W$_e'$} & \textbf{T$_e$} & \textbf{T$_H$} & \textbf{TW$_e$} & \textbf{TW$_H$} \\
\midrule
test-clean & 6.43  & 6.17  & 5.62  & 5.54  & 5.52  & \textbf{5.31} & 5.49 \\
test-other & 14.12 & 14.07 & 13.23 & 13.04 & 12.94 & 12.79 & \textbf{12.60} \\
Avg        & 10.28 & 10.12 & 9.43  & 9.29  & 9.23  & 9.05  & 9.05 \\
$\Delta$   & --    & 1.5   & 8.3   & 9.6   & 10.2  & 11.9  & \textbf{12.0} \\
\bottomrule
\end{tabular}\vspace{-2mm}
\end{table}

%% file: tables/bypass_prob.tex

\begin{table}[t]
\caption{Mean wildcard-bypass probability pooled over 9 RESPIN languages (test set). Columns 1--3 split the disputed characters by no. of dissenting validators (13.6k/3.5k/390 units). Char.: 1{,}014k agree / 17.5k disputed; Word: 174k agree / 18.9k disputed.}
\label{tab:disagree}
\centering
\scriptsize
\setlength{\tabcolsep}{3pt}
\begin{tabular}{lrrrrrrr}
\toprule
& \multicolumn{5}{c}{\textbf{Character}} & \multicolumn{2}{c}{\textbf{Word}} \\
\cmidrule(lr){2-6}\cmidrule(lr){7-8}
& Agree & Disp. & 1 & 2 & 3 & Agree & Disp. \\
\midrule
CTC        & 0.020 & 0.221 & 0.182 & 0.351 & 0.399 & 0.005 & 0.020 \\
\OTCw{}    & 0.020 & 0.234 & 0.194 & 0.369 & 0.419 & 0.005 & 0.021 \\
\OTCwt{}   & 0.024 & 0.230 & 0.193 & 0.359 & 0.397 & 0.013 & 0.039 \\
\OTCt{}    & 0.021 & 0.281 & 0.237 & 0.431 & 0.470 & 0.006 & 0.030 \\
\OTCr{}    & 0.022 & \textbf{0.302} & 0.257 & 0.454 & 0.488 & 0.006 & 0.034 \\
\OTCtw{}   & 0.023 & 0.279 & 0.235 & 0.428 & 0.462 & 0.011 & \textbf{0.041} \\
\OTCtwr{}  & 0.022 & 0.301 & 0.257 & 0.451 & 0.479 & 0.007 & 0.035 \\
\bottomrule
\end{tabular}
\end{table}